\documentclass[conference]{IEEEtran}
\IEEEoverridecommandlockouts
\usepackage{amsmath,amssymb,amsfonts}
\usepackage{bm}
\usepackage{booktabs}
\usepackage{multirow}
\usepackage{tikz}
\usepackage{svg}
\usepackage{pgfplots}
\usepgfplotslibrary{external}
\pgfplotsset{compat=1.18} 
\usepackage{cite}
\usepackage{hyperref}

\def\BibTeX{{\rm B\kern-.05em{\sc i\kern-.025em b}\kern-.08em
    T\kern-.1667em\lower.7ex\hbox{E}\kern-.125emX}}

\begin{document}

\title{
    Two-Stage Multi-task Self-Supervised Learning for Medical Image Segmentation\\
}

\author{
    \IEEEauthorblockN{Binyan Hu and A. K. Qin}
    \IEEEauthorblockA{
         \textit{Department of Computing Technologies, Swinburne University of Technology, Hawthorn, Victoria 3122, Australia}
    }
}

\maketitle

\begin{abstract}
Medical image segmentation has been significantly advanced by deep learning (DL) techniques, though the data scarcity inherent in medical applications poses a great challenge to DL-based segmentation methods. Self-supervised learning offers a solution by creating auxiliary learning tasks from the available dataset and then leveraging the knowledge acquired from solving auxiliary tasks to help better solve the target segmentation task. Different auxiliary tasks may have different properties and thus can help the target task to different extents. It is desired to leverage their complementary advantages to enhance the overall assistance to the target task. To achieve this, existing methods often adopt a joint training paradigm, which co-solves segmentation and auxiliary tasks by integrating their losses or intermediate gradients. However, direct coupling of losses or intermediate gradients risks undesirable interference because the knowledge acquired from solving each auxiliary task at every training step may not always benefit the target task. To address this issue, we propose a two-stage training approach. In the first stage, the target segmentation task will be independently co-solved with each auxiliary task in both joint training and pre-training modes, with the better model selected via validation performance. In the second stage, the models obtained with respect to each auxiliary task are converted into a single model using an ensemble knowledge distillation method. Our approach allows for making best use of each auxiliary task to create multiple elite segmentation models and then combine them into an even more powerful model. We employed five auxiliary tasks of different proprieties in our approach and applied it to train the U-Net model on an X-ray pneumothorax segmentation dataset. Experimental results demonstrate the superiority of our approach over several existing methods.

\end{abstract}


\section{Introduction}

Medical image segmentation (MIS) aims to pixel-wise delineate the targets of interest in medical images, serving as the foundational step in numerous medical image analysis pipelines. Automatic MIS has received much attention, with deep learning (DL) showcasing remarkable success on a broad spectrum of tasks \cite{isensee2021nnu}. The success of DL typically thrives on abundant annotated data. However, because of high annotation costs in the medical domain, data scarcity poses a significant challenge that limits model performance. An MIS dataset often contains as few as hundreds or tens of data samples. Supervised by such a small dataset, the trained model risks learning limited knowledge and overfitting training data, leading to poor generalisation performance when applied to real-world applications.

Self-supervised learning  (SSL) provides a promising way to tackle data scarcity by creating auxiliary tasks to facilitate solving the segmentation task. Through solving auxiliary tasks, the extra knowledge acquired can be transferred to the segmentation model to enrich its knowledge representation and thus boost generalisation performance.
Without additional annotation efforts, auxiliary tasks can be created based on the segmentation dataset in various ways. First, they can be created solely based on the images, which can be realised by many unsupervised tasks, e.g., auto-encoding and contrastive learning. Second, auxiliary tasks can be created based on the segmentation annotations, yielding tasks such as surface distance map prediction \cite{tan2018deep} and contour prediction \cite{chen2016dcan}, which sometimes can be regarded as alternative formulations of the segmentation task. Different auxiliary tasks with different mechanisms can transfer different knowledge to boost performance from distinct aspects. It is thus desired to combine their complementary advantages by involving multiple auxiliary tasks to further assist the target segmentation task.

To leverage multiple auxiliary tasks, existing approaches \cite{du2018adapting,lin2019adaptive,shi2020auxiliary} commonly adopt a multi-task learning formulation and a joint training paradigm, where the segmentation and auxiliary tasks are solved simultaneously via a certain integration of their respective losses or intermediate gradients at each iteration step.
While such a paradigm often requires homogeneous tasks with highly related learning objectives to enable effective knowledge transfer, auxiliary tasks can be quite heterogeneous in their task formulations and thus the learned knowledge.
Consequently, knowledge acquired from solving different auxiliary tasks at every training step may interfere with each other and not always benefit the target task. Specifically, the flaws are twofold. First, potential negative transfer from certain auxiliary tasks can hamper solving the segmentation task, which becomes severe with more auxiliary tasks involved. Second, for certain auxiliary tasks, interference from other tasks may impede learning valuable task-specific knowledge, making joint training infeasible to leverage these tasks.
As complementary to joint training, a more suitable way to use these auxiliary tasks is to pre-train the model with an auxiliary task and fine-tune it on the segmentation task. As will be demonstrated in \autoref{sec:impact_tf}, different auxiliary tasks have distinct choices in the appropriate training mode to enable effective knowledge transfer and otherwise yield suboptimal performance.
In a nutshell, facing heterogeneous auxiliary tasks, the interference among tasks should be alleviated and both joint training and pre-training modes need to be supported to take full advantage of auxiliary tasks.

To this end, we propose to harness multiple auxiliary tasks with a two-stage training method.
In the first stage, the target segmentation task will be independently co-solved with each auxiliary task in both joint training and pre-training modes, and the better model is selected via validation performance. This results in multiple elite segmentation models, each containing distinct knowledge transferred from an auxiliary task with the appropriate training mode. In the second stage, we employ an ensemble knowledge distillation method \cite{hinton2015distilling} to convert the models obtained with respect to each auxiliary task into a single model. As such, the diverse knowledge obtained from various auxiliary tasks is combined into an even more powerful model for inference.

We employ five representative auxiliary tasks in our method, including three unsupervised tasks, i.e., Rubik's Cube \cite{zhuang2019self}, MoCo-v3 \cite{chen2021empirical}, and VICReg \cite{bardes2021vicreg} and two surface distance prediction tasks created based on segmentation annotations. The implemented method is used to train U-Net \cite{ronneberger2015u} models on an X-ray pneumothorax segmentation dataset (SIIM-ACR Pneumothorax Segmentation Challenge) \cite{siim-acr-pneumothorax-segmentation}. Experimental results demonstrate the superiority of the proposed method in leveraging multiple auxiliary tasks to boost segmentation performance.

\section{Related Work}

\subsection{Medical Image Segmentation}
Medical image segmentation (MIS) aims at accurately delineating the target of interest (e.g., organs and lesions) in medical images, e.g., X-ray, CT, and MRI.
So far, MIS has been significantly advanced by DL, demonstrated by state-of-the-art performance in a broad spectrum of tasks.
Due to their over-parameterised nature, models in DL are data-hungry and typically require abundant annotated data to ensure good performance. Because the annotating process is expert-demanding and labour-intensive, MIS annotations are expensive and rare. As a result, DL learning with small annotated datasets has been in active research.
So far, many endeavours have been focusing on improving model architecture \cite{oktay2018attention,zhou2019unet++,cao2022swin} to more efficiently extract generalisable features.
Other works have explored learning in certain settings with no or cheaper annotations. For instance, semi-supervised learning leverages abundant unlabelled imaging data to facilitate learning. Self-supervised learning \cite{zhou2021models} can train models solely with images in an unsupervised manner. Weakly supervised learning enables learning from weak but cheap annotations. Our work focuses on the fully supervised setting where only an annotated dataset is available, and there are no additional data or annotations.

\subsection{Self-Supervised Learning}
Self-supervised learning (SSL) generally aims at feature representation learning from unlabelled data. In a common SSL paradigm, a model is pre-trained on large unlabeled data with carefully designed pretext tasks to serve as a general initialisation. It is expected that the knowledge learnt for solving the pretext tasks can be transferred to help solve various downstream tasks.
Recent advances in SSL have achieved tremendous success in training DL-based models. So far, SSL has not only been widely demonstrated as an effective tool to combat data scarcity and boost performance but has also become a foundation technique for training large models \cite{chen2021empirical}. The art of SSL is to design pretext tasks considering both the data domain and the downstream task property so that rich and useful knowledge can be extracted from the unlabeled upstream data to benefit downstream task performance. With abundant unlabelled data existing in the medical field, SSL for MIS has also been actively explored, most of which adapts the natural image counterparts to the medical image domain \cite{zhuang2019self,zhou2019models}. The prosperity of SSL provides diverse unsupervised tasks that can serve as auxiliary tasks in our supervised setting.


\subsection{Multi-Task Learning}
Multi-task learning (MTL) is a machine learning paradigm that learns multiple related tasks simultaneously.
MTL can enrich overall feature representation and boost task performance by sharing complementary knowledge obtained from solving different tasks or letting tasks act as regularisers of one another \cite{vandenhende2021multi}. MTL techniques have been extensively explored by improving model architecture or optimisation process, aiming to better capture shared information, avoid conflicts among tasks, or balance the performance of different tasks. While MTL typically relies on a dataset with multiple annotations, model training with a single-task dataset can still take advantage of MTL by involving auxiliary tasks. This setting, sometimes known as auxiliary learning, is a special case of MTL, which focuses on only the performance of the target task.
Existing auxiliary learning approaches commonly adopt the MTL formulation. Based on the joint training paradigm, different ways to adaptively adjust auxiliary task contributions have been explored. Most methods focus on task-level adaptiveness via selecting auxiliary tasks to learn from \cite{du2018adapting,kung2021efficient} or reweighting the contribution of different tasks in loss \cite{shi2020auxiliary,shamsian2023auxiliary} or gradient aggregation \cite{du2018adapting}. To achieve this, some methods rely on heuristic rules to assess task relationships \cite{du2018adapting,kung2021efficient}, and others are driven by feedback from validation set \cite{shi2020auxiliary,shamsian2023auxiliary}.

\subsection{Knowledge Distillation}
Knowledge distillation (KD) \cite{hinton2015distilling} is a popular model training technique in DL to enable knowledge transfer among models. KD was initially proposed for model compression \cite{hinton2015distilling,sau2016deep}, where a large-sized pre-trained “teacher” model trains a small-sized “student” model by supervising the student with the teacher's predictions. Through soft probabilistic labels, rich information can be transferred to the student model to facilitate its learning process \cite{bagherinezhad2018label,cheng2020explaining}.
KD techniques have received active exploration due to the flexibility to transfer knowledge solely based on model outputs, posing no restriction on the architecture or the number of teacher models. So far, KD operations have been further developed to enable distilling knowledge with different representations, such as intermittent feature maps and inter-class relationships. The utility of KD has also been extended to transfer knowledge from multiple teacher models \cite{shen2019meal}, a teacher trained together with the student \cite{zhang2018deep}, or even the student itself \cite{zhang2019your}.

\section{Method}

\subsection{Preliminary}
For conciseness, we use single-class segmentation on 2D single-channel images to derive the notations and formulations to be used in the following sections. Note that our method is naturally compatible with 3D, multi-channel, multi-modalities, and multi-class segmentation settings. We use $\bm{x}\in\mathbb{R}^{H\times W}$ to denote an image, where $H$ and $W$ represent its height and width, respectively. The binary segmentation mask annotation for each image is denoted as $\bm{y}\in{\{0,1\}}^{H \times W}$. A model for image segmentation can be represented as $f(\cdot;\bm{\theta})$, where $\bm{\theta}$ denotes its parameters.

In a typical training process, a model $f$ is trained by optimising its parameters $\bm{\theta}$ to solve the segmentation task represented by a training set $\mathbb{D}_{seg}^{tr}$, and the training process can be formulated as: 
\begin{equation}
    \bm{\theta}^*=\operatorname*{argmin}_{\bm{\theta}}(\mathbb{E}_{(\bm{x},\bm{y})\sim\mathbb{D}_{seg}^{tr}}\mathcal{L}_{seg}(\bm{y},f(\bm{x};\bm{\theta}))),
\end{equation}
where $\bm{\theta}^*$ denotes the optimised parameters, and $\mathcal{L}_{seg}$ represents the loss function, which is often the Dice loss in the MIS domain. After training, the generalisation performance of $f(\cdot,\bm{\theta}^*)$ is evaluated on a test set $\mathbb{D}_{seg}^{test}$ which does not overlap with $\mathbb{D}_{seg}^{tr}$.

Generally, we can denote the segmentation task as $\mathcal{T}_{seg}=\{\mathbb{D}_{seg}^{tr},\mathcal{L}_{seg}\}$. An auxiliary task $\mathcal{T}_i=\{\mathbb{D}_{i}^{tr},\mathcal{L}_{i}\}$ aims to train the model to enrich its knowledge representation in $\bm{\theta}$ by bringing in additional knowledge that cannot be learnt solely from $\mathcal{T}_{seg}$ and thus improve model’s generalisation performance. The dataset $\mathbb{D}_{i}^{tr}$ used by auxiliary task $\mathcal{T}_i$ is created based on $\mathbb{D}_{seg}^{tr}$, using either the images and labels in $\mathbb{D}_{seg}^{tr}$ or solely the images. Different tasks have different formulations and can be heterogeneous in terms of their input and output spaces. For example, in a segmentation task using whole images, $\bm{x}\in\mathbb{R}^{256\times256}$ and $\bm{y}\in\{0,1\}^{256\times256}$, wheras in a contrastive learning task using image patches, $\bm{x}\in\mathbb{R}^{128\times128}$ and  $\bm{y}\in\mathbb{R}^{1024}$. To differentiate the inputs and outputs among tasks, we use $\bm{x}_{seg}$ and $\bm{y}_{seg}$ to denote the image and label for segmentation and use $\bm{x}_{i}$ and $\bm{y}_{i}$ to denote the image and label for auxiliary task $i$. Our work involves multiple auxiliary tasks $\{\mathcal{T}_i\}_{i=1}^{N}$ to improve model training, where $N$ denotes the total number of auxiliary tasks. To enable knowledge transfer, every auxiliary task shares a certain part of the model with the segmentation task, parameterised by  $\bm{\theta}_{sh}$, and every task, including the segmentation task, has its task-specific head. We use $\bm{\theta}_{seg}$ to denote the segmentation-specific parameters, and $\bm{\theta}_{i}$ to denote the parameters specific to auxiliary task $\mathcal{T}_i$.

\subsection{Framework}
\label{sec:framework}

\begin{figure*}[ht]
    \centering
    \includegraphics[width=0.9\textwidth]{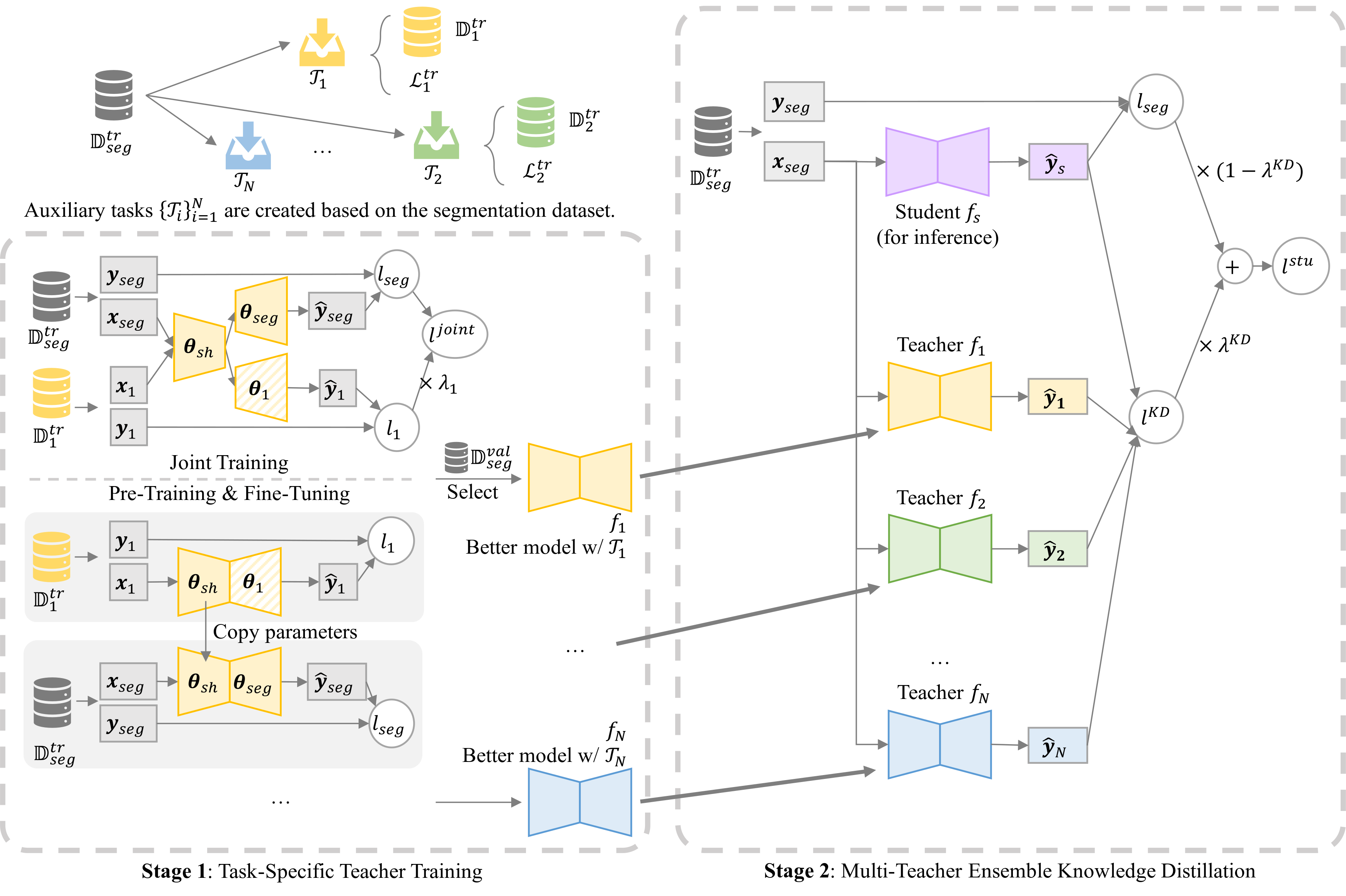}
    \caption{An illustration of the proposed method. Best viewed in colour. It leverages $N$ auxiliary tasks $\{\mathcal{T}_i\}_{i=1}^{N}$, created based on the training set $\mathbb{D}_{seg}^{tr}$, to boost segmentation performance. The method is composed of two training stages. 
    In the first stage, the target segmentation task will be independently co-solved with each auxiliary task $\mathcal{T}_i$ in both joint training and pre-training modes.
    In the joint training mode, the model is trained to concurrently solve the segmentation and auxiliary tasks through a weighted combination of their respective losses $l_{seg}$ and $l_i$.
    In the pre-training mode, the model is first trained to solve the auxiliary task with $l_i$ and then transferred to be fine-tuned on the segmentation task with $l_{seg}$.
    The better model obtained by the two modes $f_i$ is selected via the performance evaluated on the validation set $\mathbb{D}_{seg}^{val}$.
    In the second stage, the models obtained with respect to each auxiliary task (as teachers) $\{f_i\}_{i=1}^N$ are converted into a single model (as student) $f_s$ using an ensemble knowledge distillation method, which matches the student's segmentation output $\hat{\mathbf{y}}_s$ with the ensemble of the teachers' segmentation outputs $\{\hat{\mathbf{y}}_i\}_{i=1}^N$.
    Finally, the student model $f_s$ is returned for inference. }
    \label{fig:framework}
\end{figure*}

Given a target MIS dataset and multiple auxiliary tasks, the proposed method aims to obtain a model with good performance on the MIS dataset by leveraging the auxiliary tasks. As shown in \autoref{fig:framework}, the method is composed of two training stages. 
First, the target segmentation task $\mathcal{T}_{seg}$ will be independently co-solved with each auxiliary task $\mathcal{T}_i$ in both joint training and pre-training modes, and the better model obtained by the two modes $f_i$ is selected via validation performance.
As a result, multiple elite segmentation models $\{f_i\}_{i=1}^N$ are obtained, each containing useful and distinct knowledge transferred from an auxiliary task. Second, these models are treated as teacher models and are compressed into a single model $f_s$ via ensemble knowledge distillation \cite{hinton2015distilling}. In this way,  the diverse knowledge obtained from multiple auxiliary tasks is aggregated to obtain an even more powerful student model, which is output for inference. The following sections describe the two training stages in detail.

\subsection{Task-Specific Teacher Training}
\label{sec:task_specific}
This stage aims to obtain a segmentation model independently facilitated by each auxiliary task.
In this way, the interference among auxiliary tasks can be avoided, and the most appropriate training mode for each auxiliary task can be independently applied to make best use of each auxiliary task.
For auxiliary task $\mathcal{T}_i$, we train two models to co-solve $\mathcal{T}_i$ and segmentation $\mathcal{T}_{seg}$ in two modes: joint training and pre-training, respectively.

In the joint training mode, the model is trained to concurrently solve the segmentation and auxiliary tasks in a multi-task fashion. At each iteration step, we sample a batch of samples for the segmentation task and the auxiliary task, respectively, i.e., $\bm{x}_{seg},\bm{y}_{seg}\sim\mathbb{D}_{seg}^{tr}$ and $\bm{x}_{i},\bm{y}_{i}\sim\mathbb{D}_{i}^{tr}$. Then the inputs are fed to the model, and the training loss is calculated as
\begin{equation}
\begin{split}
    l^{joint}=&\mathcal{L}_{seg}(\bm{y}_{seg},f(\bm{x}_{seg};\bm{\theta}_{sh},\bm{\theta}_{seg}))\\
    &+\lambda_i\mathcal{L}_{i}(\bm{y}_{i},f(\bm{x}_{i};\bm{\theta}_{sh},\bm{\theta}_{i})),
\end{split}
\end{equation}
where $\lambda_i$ denotes a scalar coefficient to weight the loss contribution of task $i$. A larger $\lambda_{i}$ encourages the trained model to learn more from the auxiliary task and vice versa. All the model parameters $\bm{\theta}_{sh}$, $\bm{\theta}_{seg}$, and $\bm{\theta}_{i}$ are jointly updated to minimise $l^{joint}$.

The pre-training mode involves two training steps. First, the model is pre-trained to solve the auxiliary task via the task-specific loss and data:
\begin{equation}
    l_{i}=\mathcal{L}_{i}(\bm{y}_{i},f(\bm{x}_{i};\bm{\theta}_{sh},\bm{\theta}_{i})),
\end{equation}
to train $\bm{\theta}_{sh}$ and $\bm{\theta}_{i}$. Afterwards, the shared parameters $\bm{\theta}_{sh}$ are transferred to initialise the segmentation model in the fine-tuning step, where the model is trained to solve the segmentation task via
\begin{equation}
    l_{seg}=\mathcal{L}_{seg}(\bm{y}_{seg},f(\bm{x}_{seg};\bm{\theta}_{sh},\bm{\theta}_{seg})).
\end{equation}
In the fine-tuning step, the pre-trained $\bm{\theta}_{sh}$ are continuously trained along with randomly initialised $\bm{\theta}_{seg}$.

After the two models are trained, we select the better one based on the validation performance evaluated by the validation set $\mathbb{D}_{seg}^{val}$.
By selecting the most appropriate training mode for leveraging each auxiliary task, the chance of negative transfer is reduced. Also, because each auxiliary task takes effect independently, the conflicts between auxiliary tasks are avoided. Another advantage of independent training is that this step can be naturally distributed to different machines to accelerate training.

\subsection{Multi-Teacher Ensemble Knowledge Distillation}
\label{sec:multiaux_disitl}

This stage aims to compress those models obtained by the previous stage, denoted as $\{f_i\}_{i=1}^N$ to a single model $f_s$ for inference. In this way, the diverse knowledge obtained from multiple auxiliary tasks is aggregated to obtain an even more powerful model. To achieve this, we employ an ensemble knowledge distillation \cite{hinton2015distilling}. The previously obtained models are treated as teachers to distil their knowledge into the student model $f_s$. For each input image, the average of all the teacher model's probabilistic prediction maps, i.e., their ensemble prediction $\frac{1}{N}\sum_{i=1}^{N}f_i(\bm{x}_{seg})$, is used as the pseudo annotation to train the student model.
Considering that the pseudo annotation inevitably carries erroneous predictions from the teacher models, only using $l^{KD}$ to supervise the student is prone to misleading. To address this issue, following common practice, we define the loss for student training as a linear combination of the loss w.r.t. the ground-truth annotation and the KD loss w.r.t. the pseudo annotation. Putting together, the total loss for training the student model is formulated as
\begin{equation}
\begin{split}
    l^{stu}=&(1-\lambda^{KD})\mathcal{L}_{seg}(\bm{y}_{seg},f_s(\bm{x}_{seg}))\\
    &+\lambda^{KD}\mathcal{L}_{seg}(\frac{1}{N}\sum_{i=1}^{N}f_i(\bm{x}_{seg}),f_s(\bm{x}_{seg})),
\end{split}
\end{equation}
where $\lambda^{KD}$ is a coefficient balancing the impact of ground-truth annotations and pseudo annotations. We set $\lambda^{KD}=1/(N+1)$ intuitively to treat every pseudo annotation from a teacher model as equally important as the ground-truth annotation.

\section{Experiments}

We first describe the datasets used in this study in \autoref{sec:dataset}, then introduce the experimental setup in \autoref{sec:setup}. Then, in \autoref{sec:impact_tf}, we show the crucial impact of the training mode for leveraging an auxiliary task. Based on the results in this section, we determine the training mode for each auxiliary task in the implementation of our method to be used in the subsequent experiments.
In \autoref{sec:overall}, we compare our proposed method with several existing methods for leveraging auxiliary tasks to demonstrate the superiority of our method. Then in \autoref{sec:analysis_save}, we measure the impact of our method on saving training data compared to the conventional training method. Finally, we provide further analysis on the auxiliary task contributions \autoref{sec:analysis_task} and the impact of the KD loss coefficient \autoref{sec:analysis_labmda_kd}.

\subsection{Dataset}
\label{sec:dataset}

\textbf{PNE} (SIIM-ACR Pneumothorax Segmentation Challenge) \cite{siim-acr-pneumothorax-segmentation}
is the largest public dataset for pneumothorax segmentation. The images take a subset of the ChestX-ray14 dataset \cite{wang2017chestx} released by the National Institutes of Health (NIH), which consists of both positive and negative pneumothorax cases. In the positive cases, pixel-wise delineations of pneumothorax are provided. In this study, we take all 2669 samples with positive pneumothorax for experiments. In cases where multiple annotations are available, we merge the annotation masks by a union operation to include all the positive regions. All images are resized
to $256 \times 256$. We randomly split the dataset into 1601/267/801 samples for training/validation/testing.



\subsection{Experimental Setup}
\label{sec:setup}


\textbf{Common setup}. To simulate a scenario suffering data scarcity, we randomly select 200 samples for training, which is a common number of data accessible in the application. We train 2D U-Net \cite{ronneberger2015u} models with the backbone of ResNet-18 \cite{he2016deep}, which is commonly adopted for MIS tasks. To construct the U-Net architecture, we adopt group normalisation layers \cite{wu2018group} and ReLU activation layers. For data augmentation, we use random translation, zooming, rotation, Gaussian noise, Gaussian blur, brightness jittering, contrast jittering, and gamma jittering following \cite{isensee2021nnu}. We use a batch size of 16 and the Dice loss function to train every segmentation model.
We adopt the Rectified Adam (RAdam) optimiser \cite{liu2019variance} for model training, where the learning rate is initialised to 0.001 and decayed by the poly annealing scheduler with a power rate of 0.9 throughout the training process.

\textbf{Auxiliary Tasks}. We employ the following five representative auxiliary tasks, covering both segmentation annotation-based tasks (SDM-in and SDM-out) and unsupervised tasks (RKB, MoCo, and VICReg), to implement our method:
\begin{itemize}
    \item \textbf{SDM-in} and \textbf{SDM-out}: Surface distance map (SDM) is an alternative representation of segmentation mask, with a rigorous mapping between them through Euclidean distance transform. SDM prediction can help produce smooth segmentation boundaries and reduce spatially isolated errors \cite{tan2018deep}. In this study, we consider predicting two kinds of SDMs as auxiliary tasks. \textbf{SDM-in} is obtained by replacing the intensity of each pixel in the foreground with the distance to its closest background pixel and setting the background regions to 0. Conversely, \textbf{SDM-out} is obtained by replacing the intensity of each pixel in the background with the distance to its closest foreground pixel and setting the foreground regions to 0.
    \item \textbf{RKB} (Rubik's Cube) \cite{zhuang2019self} is a predictive SSL task that trains a model to predict the correct order of a set of shuffled image patches cropped from the same image.
    \item \textbf{MoCo} (Momentum Contrast-v3) \cite{chen2021empirical} is a contrastive SSL task that trains a model to differentiate the image patches cropped from different images. The task is designed to contrast the latent feature representations of different image patches so that the features of patches cropped from the same image (positive pair) are pulled together, and the features of patches cropped from different images (negative pair) are pushed apart.
    \item \textbf{VICReg} (Variance-Invariance-Covariance Regularization) \cite{bardes2021vicreg} is another contrastive SSL task. It has a similar goal as MoCo but employs a different contrastive learning formulation via a variance loss term and an invariance loss term. In addition, it promotes feature diversity by a covariance loss term.
\end{itemize}

\subsection{Impact of Training Mode to Leverage Auxiliary Tasks}
\label{sec:impact_tf}

\begin{table}
    \centering
    \caption{Comparison of Dice scores obtained by models trained to co-solve the segmentation task and one auxiliary task with different training modes of joint training and pre-training. Every experiment is run over 5 times. For each auxiliary task, the best performer in terms of the mean value is highlighted in bold. The result obtained via conventional training without any auxiliary task (the first row) is also reported as a reference.}
    \label{tab:multiaux_usage}

    \begin{tabular}{c|cc|c}
        \toprule
        \multirow{2}{*}{Aux. Task}  &  \multicolumn{2}{c|}{Training Mode} & \multirow{2}{*}{Dice Score (\%) ↑}\\
                                    &  Joint Training   &  Pre-Training                 & \\
        \midrule
        None                        & / & / &38.16±0.60\\
        \midrule
        \multirow{2}{*}{SDM-in}     & \checkmark& &\textbf{38.65±0.37}\\
                                    &  &  \checkmark& 37.56±0.61\\
        \midrule
        \multirow{2}{*}{SDM-out}    & \checkmark& &\textbf{38.36±0.23}\\
                                    & & \checkmark&37.17±0.74\\
        \midrule
        \multirow{2}{*}{RKB}        & \checkmark& &38.45±0.17\\
                                    & & \checkmark&\textbf{39.23±0.27}\\
        \midrule
        \multirow{2}{*}{MoCo}       & \checkmark& &38.13±0.56\\
                                    & & \checkmark&\textbf{39.23±0.42}\\
        \midrule
        \multirow{2}{*}{VICReg}     &  \checkmark&  & 38.47±0.33\\
                                    & & \checkmark&\textbf{38.67±0.67}\\
        \bottomrule
    \end{tabular}
\end{table}

\begin{table*}[ht]
    \centering
    \caption{Overall comparison of the proposed method with existing methods leveraging auxiliary tasks. The results of using a single auxiliary task with joint training or pre-training modes are also presented for reference. All the methods are used for model training with 200 training samples. Every experiment is run 5 times, and the Dice scores (mean$\pm$std) are reported. The best performer in terms of mean value is highlighted in bold, and the second best is underlined.}
    \label{tab:multiaux_overall}

    \begin{tabular}{cl|ccccc|c}
    \toprule
        \multicolumn{2}{c|}{\multirow{2}{*}{
            Training Method}}   &\multicolumn{5}{c|}{Aux. Tasks}            & \multirow{2}{*}{Dice Score (\%) ↑}\\
        \multicolumn{2}{c|}{}   &  SDM-in&  SDM-out&  MoCo&  RKB&  VICReg   &   \\
    \midrule
        \multicolumn{2}{c|}{Conventional}& & & & & &38.16±0.60\\
    \midrule
        \multirow{5}{*}{Single Aux. Task}& \multirow{5}{*}{Joint Training / Pre-Training} &  \checkmark&  &  &  &  & 38.65±0.37 / 37.56±0.61\\
        &  &  &  \checkmark&  &  &  & 38.36±0.23 / 37.17±0.74\\
        &  &  &  &  \checkmark&  &  & 38.13±0.56 / 39.23±0.42\\
        &  &  &  &  &  \checkmark&  & 38.45±0.17 / 39.23±0.27\\
        &  &  &  &  &  &  \checkmark& 38.47±0.33 / 38.67±0.67\\
    \midrule
        \multirow{7}{*}{Multi Aux. Tasks}& Joint Training& \checkmark& \checkmark& \checkmark& \checkmark& \checkmark&38.47±0.36\\
        & Multi-Task Pre-Training& \checkmark& \checkmark& \checkmark& \checkmark& \checkmark&38.35±0.37 \\
        & Ensemble  & \checkmark& \checkmark& \checkmark& \checkmark& \checkmark& \underline{40.33±0.34} \\
        & GCS \cite{du2018adapting}& \checkmark& \checkmark& \checkmark& \checkmark& \checkmark&38.55±0.42\\
        & PCGrad \cite{yu2020gradient}& \checkmark& \checkmark& \checkmark& \checkmark& \checkmark&38.90±0.23\\
        & OL-AUX \cite{lin2019adaptive}& \checkmark& \checkmark& \checkmark& \checkmark& \checkmark&38.10±0.33\\
        & AMAL\cite{sivasubramanian2023adaptive}& \checkmark& \checkmark& \checkmark& \checkmark& \checkmark&38.60±0.24\\
        & ours& \checkmark& \checkmark& \checkmark& \checkmark& \checkmark&\textbf{42.05±0.20}\\
    \bottomrule
    \end{tabular}
\end{table*}

\begin{table}[ht]
    \caption{Segmentation performance obtained with different numbers of training data with the conventional training method. Every experiment is run 5 times, and the Dice scores (mean$\pm$std) are reported.}
    \label{tab:multiaux_n_data}
    \centering
    \begin{tabular}{c|c}
    \toprule
        \# Training Data& Dice Score (\%) ↑\\
    \midrule
        200& 38.16±0.60\\
        400& 41.27±0.14\\
        500& 42.42±0.38\\
        600& 43.46±0.61\\
        800& 45.14±0.10\\
        1600&47.71±0.23\\
    \bottomrule
    \end{tabular}
\end{table}

For each auxiliary task, we independently use it to conduct two experiments, each using the auxiliary task to facilitate training a segmentation model with either joint training or pre-training following the respective procedures described in \autoref{sec:task_specific}. The experimental results are reported in \autoref{tab:multiaux_usage}. It can be observed that different auxiliary tasks work well with different training modes. SDM-in and SDM-out work well only via joint training while drastically degrading the segmentation performance with pre-training. By contrast, RKB, MoCo, and VICReg work better with pre-training while offering no or limited improvements when using joint training. These results suggest the appropriate training mode differs for each auxiliary task, and no single training mode works well for all the auxiliary tasks.
Furthermore, to take full advantage of different auxiliary tasks, both joint training and pre-training modes need to be supported, which motivates the design of our proposed method.


\subsection{Overall Comparison}
\label{sec:overall}

This section compares the proposed method with several existing methods that leverage multiple auxiliary tasks to boost model performance. We compare the following methods, including intuitive methods for multi-task learning or pre-training, and existing methods proposed for leveraging multiple auxiliary tasks to boost model performance under scarce training data:
\begin{itemize}
    \item \textbf{Joint Training} is the most commonly used way to leverage auxiliary tasks.
    The segmentation task and the auxiliary tasks are used together to train the model via a weighted combination of their respective losses. Finally, the segmentation model is returned.
    \item \textbf{Multi-Task Pre-Training} is a common way to perform model unsupervised pre-training with multi-task SSL, which is adapted to our supervised setting. It first pre-trains a backbone model on all the auxiliary tasks using Joint Training. Then, the pre-trained backbone is transferred to initialise the target segmentation model, which is fine-tuned on the segmentation dataset.
    \item \textbf{Ensemble} is an intuitive way to aggregate the knowledge learnt from multiple auxiliary tasks. Similar to the first stage of the proposed method, it first uses each auxiliary task independently to facilitate the training of a segmentation model. Then, it directly uses the ensemble of all the obtained models for inference. 
    \item \textbf{GCS} (Gradient Cosine Similarity) \cite{du2018adapting} is an existing method that utilizes the similarity between the gradients from each auxiliary task and those from the primary task to determine whether to learn from each auxiliary task at every iteration step.
    \item \textbf{PCGrad} (Projecting Conflicting Gradients) \cite{yu2020gradient} is an MTL method adapted to our setting. At every iteration step, for each auxiliary task, the gradients that conflict with the segmentation task are removed.
    \item  \textbf{OL-AUX} (Online Learning for Auxiliary losses) \cite{lin2019adaptive} adaptively changes the loss weight based on the gradient inner product w.r.t the main task to decrease the long-term value of the main task loss.
    \item \textbf{AMAL} (Adaptive mixing of Auxiliary Losses) \cite{sivasubramanian2023adaptive} adaptively changes the loss weight based on gradient feedback from the validation set.
\end{itemize}


The results are reported in \autoref{tab:multiaux_overall}. It can be seen that existing methods based on the joint training paradigm (all except for Multi-Task Pre-Training, Ensemble, and ours) can hardly integrate the advantages of multiple auxiliary tasks, evidenced by their equal or marginally better performance compared to the baselines of conventional training and joint training with a single auxiliary task. Among these methods, only PCGrad, which removes the gradients hampering solving the segmentation task can surpass all these baselines. This implies a severe interference among the learning heterogeneous tasks, which cannot be well tackled by certain designs trying to filter out negative transfer (GCS and PCGrad) or adaptively integrate the contributions of different auxiliary tasks (AMAL) within the joint training paradigm. Also, these compared methods cannot match the performance obtained by pre-training with a single auxiliary task of MoCo or RKB, showing the shortcoming of joint training to leverage these tasks. By contrast, those methods that determine the training mode for each auxiliary task (Ensemble and ours) significantly improve the performance over all the baselines and compared methods. This demonstrates the necessity to adaptively determine the training mode for different auxiliary tasks and the superiority of using each auxiliary task independently to avoid interference. In addition, our method outperforms Ensemble with a single model, indicating the effectiveness of using ensemble knowledge distillation to aggregate the knowledge from various auxiliary tasks.

\subsection{Analysis on Saving Training Data}
\label{sec:analysis_save}

We probe the number of training data that can be saved by our method. We train segmentation models using conventional training with different numbers of data, ranging from 200 to 1600 and report the corresponding segmentation performance in \autoref{tab:multiaux_n_data}. It can be seen that the performance of our method with only 200 training samples matches that obtained with 400 - 500 training samples with conventional training, indicating that our method can effectively boost the segmentation performance when the training data is limited.

\subsection{Analysis on Auxiliary Task Contribution}
\label{sec:analysis_task}

\begin{table}
    \caption{Dice scores obtained by the proposed method employing different combinations of auxiliary tasks. Every experiment is run 5 times. The best performer and those not significantly different from the best ($p\ge0.05$), are highlighted in bold.}
    \label{tab:multiaux_ablate_task}
    \centering
    \resizebox{\linewidth}{!}{\begin{tabular}{c|ccccc|c}
    \toprule
        &\multicolumn{5}{c}{Aux. Tasks Involved}& Dice Score (\%) ↑\\
        \# Tasks&MoCo&  RKB&  SDM-in&  SDM-out&  VICReg& \\
    \midrule
        5&\checkmark&  \checkmark&  \checkmark&  \checkmark&  \checkmark& \textbf{42.05}\\
    \midrule
        4&&  \checkmark&  \checkmark&  \checkmark&  \checkmark& 41.78\\
        &\checkmark&  &  \checkmark&  \checkmark&  \checkmark& 41.57\\
        &\checkmark&  \checkmark&  &  \checkmark&  \checkmark& \textbf{42.00}\\
        &\checkmark&  \checkmark&  \checkmark&  &  \checkmark& \textbf{41.90}\\
        &\checkmark&  \checkmark&  \checkmark&  \checkmark&  & \textbf{42.12}\\
    \midrule
        3&&  \checkmark&  \checkmark&  \checkmark&  & 41.81\\
        &\checkmark&  &  \checkmark&  \checkmark&  & 41.60\\
        &\checkmark&  \checkmark&  &  \checkmark&  & \textbf{42.10}\\
        &\checkmark&  \checkmark&  \checkmark&  &  & \textbf{42.10}\\
    \midrule
        2&&  \checkmark&  \checkmark&  &  & \textbf{41.96}\\
        &\checkmark&  &  \checkmark&  &  & 41.77\\
        &\checkmark&  \checkmark&  &  &  & \textbf{42.04}\\
    \midrule
        1&\checkmark&  &  &  &  & 41.16\\
        &&  \checkmark&  &  &  & 41.68\\
        &&  &  \checkmark&  &  & 40.18\\
        &&  &  &  \checkmark&  & 40.27\\
        &&  &  &  &  \checkmark& 41.06\\
    \bottomrule
    \end{tabular}}
\end{table}

As shown in \autoref{tab:multiaux_ablate_task}, we start by using all 5 auxiliary tasks as a baseline and ablate each auxiliary task one by one to obtain the results with 4 auxiliary tasks. Then the best performer with 4 auxiliary tasks is set as the new baseline to further ablate an auxiliary task to conduct the experiments involving 3 auxiliary tasks. This process is repeated until only 2 auxiliary tasks are left. Finally, the results obtained using a single auxiliary task in our method are listed for reference. It can be observed from the results that the performance obtained using all 5 tasks is among the best performers, whereas using less than 5 auxiliary tasks cannot guarantee optimal performance. Besides, no single auxiliary task alone can obtain good performance. These results indicate the effectiveness of using all the auxiliary tasks to integrate their complementary advantages and ensure good performance. It is also shown that certain auxiliary tasks (RKB and MoCo in our case) play more important roles than the others, evidenced by those inferior results obtained without these two tasks.

\subsection{Analysis on KD loss coefficient}
\label{sec:analysis_labmda_kd}

\begin{table}[ht]
    \caption{Dice scores (mean$\pm$std) obtained with different $\lambda^{KD}$. Every experiment is run 5 times. The best performer in terms of mean value is shown in bold, and the second best is underlined.}
    \label{tab:lambda_kd}
    \centering
    \begin{tabular}{c|c}
    \toprule
        $\lambda^{KD}$& Dice Score (\%) ↑\\
    \midrule
        0& 36.61±0.76\\
        0.1& 37.88±0.47\\
        0.2& 38.33±0.28\\
        0.3& 39.37±0.44\\
        0.4& 40.45±0.13\\
        0.5& 41.50±0.26\\
        0.6& 41.95±0.26\\
        0.7& \textbf{42.07±0.10}\\
        0.83 (ours)& \underline{42.05±0.20}\\
        0.9& 41.95±0.26\\
        1&41.64±0.12\\
    \bottomrule
    \end{tabular}
\end{table}

We analyse the impact of the coefficient in the KD loss $\lambda^{KD}$ by setting it to certain values within $[0,1]$.  $\lambda^{KD}=0$ corresponds to only using the segmentation loss for training, and $\lambda^{KD}=1$ corresponds to only using the distillation loss. Recall that our method is implemented by setting $\lambda^{KD}=1/(N+1)=0.83$. As shown in \autoref{tab:lambda_kd}, the best result is obtained at $\lambda^{KD}=0.7$, and our setting achieves the second best. The inferior performance obtained by a $\lambda^{KD}$ of 0 or 1 indicates the importance of incorporating the supervision from both the ground-truth annotations and pseudo annotations for good performance. The results also show that the KD method in Stage 2 is not sensitive to the choice of $\lambda^{KD}$ as values within $[0.6,0.9]$ yield similar results.

\section{Conclusions and Future Work}
We proposed a two-stage multi-task self-supervised learning approach to address the issue of data scarcity in medical image segmentation. Specifically, given the target segmentation task and the available training set for solving the task, multiple different auxiliary tasks are created in a self-supervised mode. After that, the target task is first independently co-solved with each auxiliary task in both joint training and pre-training modes, with the better model (based on validation performance) generated by the two training modes selected. Then, the models obtained with respect to each auxiliary task are used to produce a single model using an ensemble knowledge distillation method. This two-stage learning approach allows various auxiliary tasks to be more effectively leveraged to help solve the target segmentation task. Experiments validated the performance superiority of our approach compared to several existing methods that utilize multiple auxiliary tasks.

In the future, we plan to investigate ways to reduce the computation cost incurred by the model training with every auxiliary task in two modes. We will also study how to enable more effective and efficient knowledge transfer from auxiliary tasks, where evolutionary multi-task optimization-based approaches \cite{song2019multitasking,wu2022evolutionary} can be a potential solution. Further, we will explore the relationships between the models generated in auxiliary tasks, e.g., via graph matching \cite{gong2016discrete} and learning vector quantization \cite{qin2005initialization} techniques, to increase the diversity of auxiliary tasks for augmenting their helpfulness to the target task.



\bibliographystyle{ieeetr}
\bibliography{refers}

\end{document}